\documentclass[sigconf, anonymous=false, nonacm]{acmart}
\usepackage[utf8]{inputenc}
\usepackage{hyperref}
\usepackage{graphicx}
\usepackage{subcaption}
\usepackage{xspace}
\usepackage[toc,page]{appendix}
\usepackage[parfill]{parskip}

\title{Face Pasting Attack}

\author{Niklas Bunzel}
\affiliation{%
	\institution{Fraunhofer SIT / ATHENE}
	\city{Darmstadt}
	\country{Germany}
}
\email{bunzel@sit.fraunhofer.de}

\author{Lukas Graner}
\affiliation{%
	\institution{Fraunhofer SIT / ATHENE}
	\city{Darmstadt}
	\country{Germany}
}
\email{graner@sit.fraunhofer.de}

\newcommand{\conf}{\emph{confidence}\xspace}
\newcommand{\stealth}{\emph{stealthiness}\xspace}

\begin{abstract}
	Cujo AI and Adversa AI hosted the MLSec face recognition challenge. The goal was to attack a black box face recognition model with targeted attacks. The model returned the \conf of the target class and a \stealth score. For an attack to be considered successful the target class has to have the highest \conf among all classes and the \stealth has to be at least $0.5$. In our approach we paste the face of a target into a source image. By utilizing position, scaling, rotation and transparency attributes we reached 3rd place. Our approach took approximately $200$ queries per attack for the final highest score and about $\sim$$7.7$ queries minimum for a successful attack. The code is available at \url{https://github.com/bunni90/FacePastingAttack}.
\end{abstract}


\date{}

\begin{document}
\maketitle

\section{Introduction}
\label{sec:intro}
The Machine Learning Security Evasion Competition takes place at the DEFCON AI Village since 2019 %
with tasks like attacking ML based malware detection or antiphishing systems or defending against attacks. In this years competition there are two tasks evading a anti phishing system and evading a face recognition system. In this paper we focus on the task of evading the face recognition system. Face recognition is being used in production systems more and more, for example in video identification\footnote{\url{https://www.webid-telecom.de/de/\#webidai}} or to unlock and pay with your smartphone. But neural networks are prone to adversarial attacks and a variety of diverse attacks exist. In this challenge a black box face recognition system, which is secured against adversarial attacks is to be attacked by impersonation. This challenge and the attacks implemented therefore closely reflect a real world attack scenario.
\section{Related Work}
\label{sec:related}
The first adversarial attack L-BFGS~\cite{SzegedyZSBEGF13} was published in 2014, since then a lot of research was conducted in this field. Madry et. al implemented the PGD~\cite{MadryMSTV18} attack which is still considered as one of the strongest white box attacks. For black box attacks two main approaches have emerged: transfer-based attacks~\cite{PapernotMGJCS17} and query-based attacks~\cite{ChenZSYH17}.\\
Brown et. al developed the adversarial patch~\cite{advpatch}, where the adversarial perturbation size and shape are constrained. The adversarial patch can be placed into an image to force a specific classification.
In~\cite{JoshiMSH19} Joshi et. al investigate semantic attacks which may not necessarily add imperceptible noise, but rather manipulate semantic features in a way that they look more natural.
Kaziakhmedov et. al attack MTCNN face detection with printable stickers (placed on face cheeks) in~\cite{Kaziakhmedov_2019}.
Komkov et. al perturbing a plane on a hat in order to fool ArcFace face recognition system~\cite{advhat}.
In~\cite{Sharif19AGNs} Sharif et. al are implementing a perturbation framework and demonstrate it by building adversarially imprinted glasses to mislead face recognition systems. %
In~\cite{TransferableFaceRecognition} the authors implemented a network to create transferable adversarial attacks on face recognition models.
Guo et. al utilize meaningful stickers to adversarialy attack face recognition systems~\cite{meaningfulPatch}.

\section{Background}
\label{sec:background}
\paragraph{The challenge}
The MLSec face recognition challenge 2022, held from August 12 to September 23, is part of the DEFCON AI Village and is organized by Adversa AI, Cujo AI and Robust Intelligence. 
The goal is to attack a given face recognition system, by slightly altering given face images in order to be misclassified.
Since access to the model is provided solely through API calls without any further information given about it, such as architecture or training data, the model is effectively a black box. 

After passing an (altered) image, two scores are returned by the API: 1. \conf $\in [0;1]$, which represents the target class probability score of the image calculated by the model, and 2. \stealth $\in [0;1]$, which is a similarity score (presumably based on SSIM) to the unaltered source images. 

For the challenge 10 source images are given $I_0,..., I_9$, depicting the faces of 10 persons which represent the set of classes $C_0,..., C_9$. The goal is for each source image $I_{s}$ to construct 9 alterations $I^{t}_{s}$ (where $t \neq s$), that will be classified as a different class $C_{t}$ by the face recognition system. This results in 90 attacks in total, comprising all unequal pairs of source and target classes.

Although contestants compete for the highest \conf only, an attack is only considered successful, when the probability score of the target class (i.e. \conf) is the highest among all classes and \stealth is at least $0.5$.

\section{Face Pasting Attack}
\label{sec:attack}
\begin{figure}
	\begin{subfigure}{\linewidth}
		\label{fig:faces}
		\includegraphics[width=0.19\linewidth]{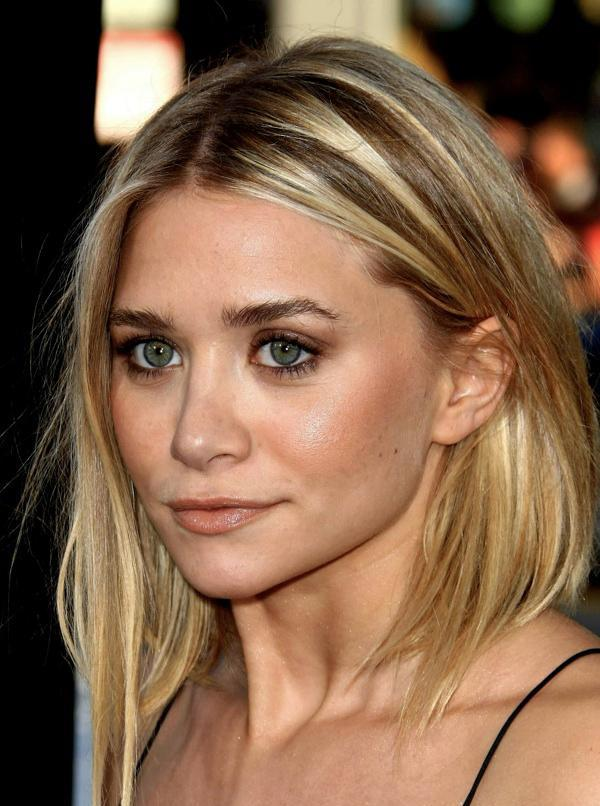}
		\includegraphics[width=0.19\linewidth]{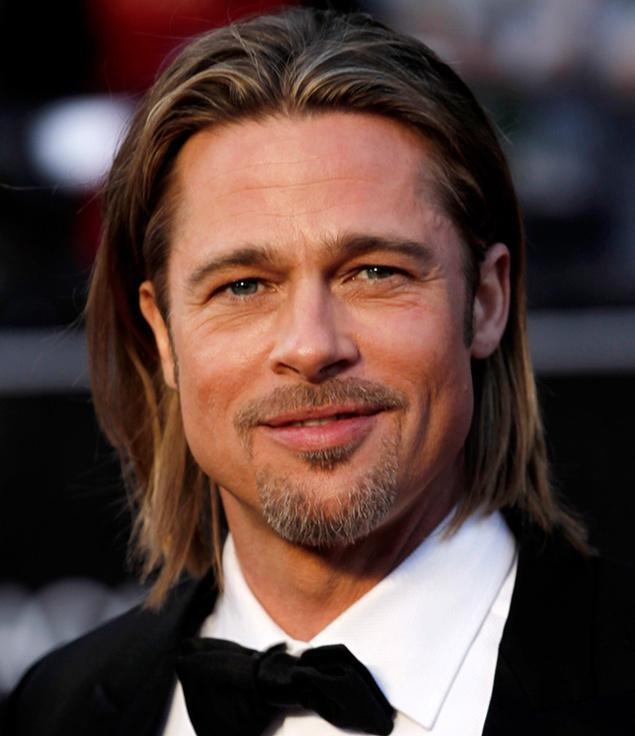}
		\includegraphics[width=0.19\linewidth]{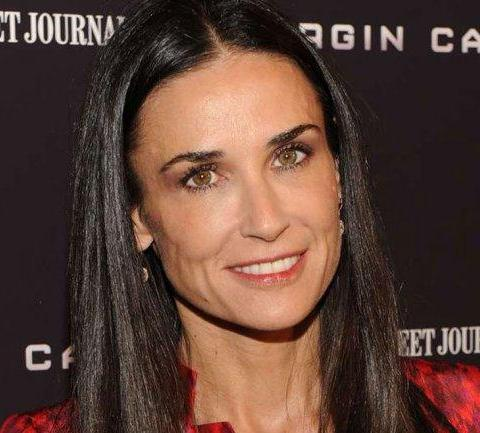}
		\includegraphics[width=0.19\linewidth]{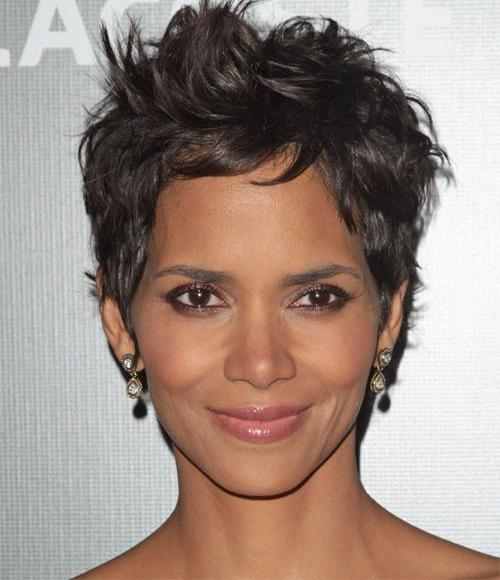}
		\includegraphics[width=0.19\linewidth]{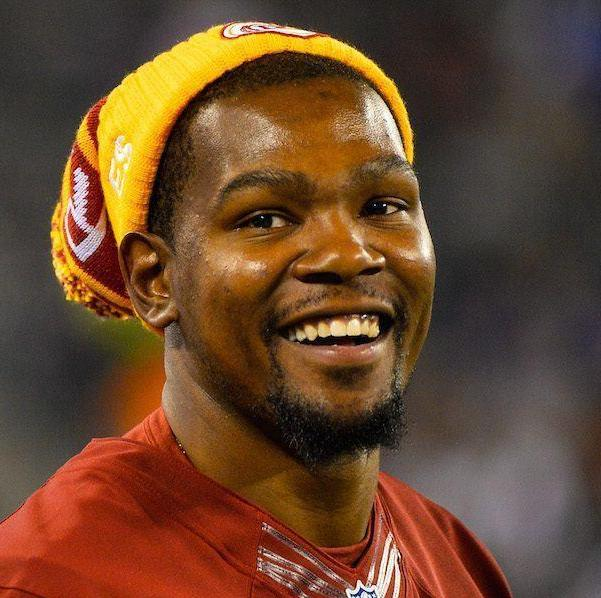}
	\end{subfigure}%
	\vskip\baselineskip
	\begin{subfigure}{\linewidth}
		\label{fig:man_masks}
		\includegraphics[width=0.19\linewidth]{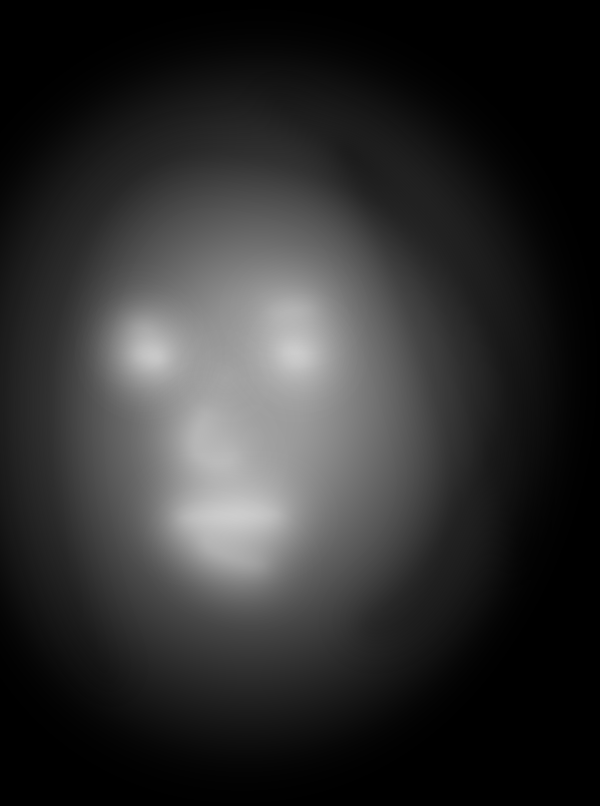}
		\includegraphics[width=0.19\linewidth]{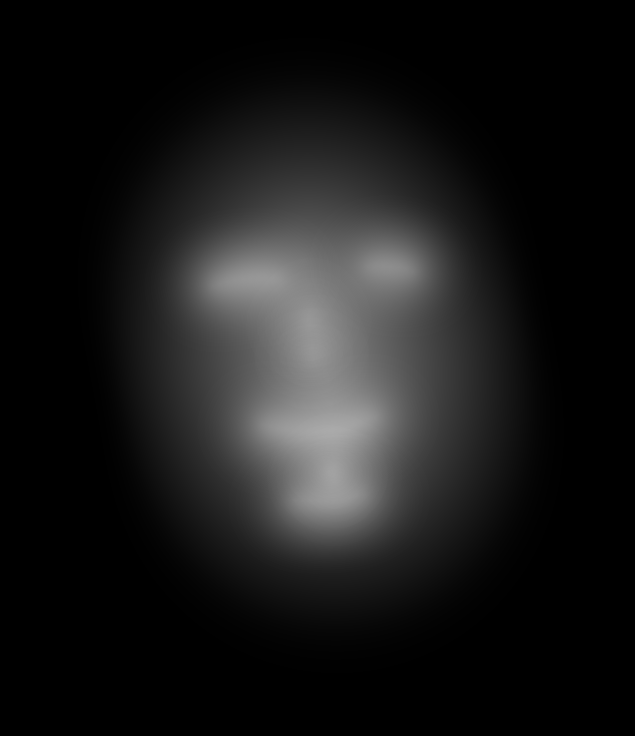}
		\includegraphics[width=0.19\linewidth]{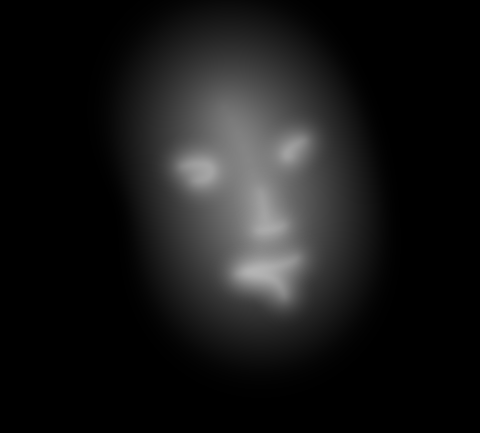}
		\includegraphics[width=0.19\linewidth]{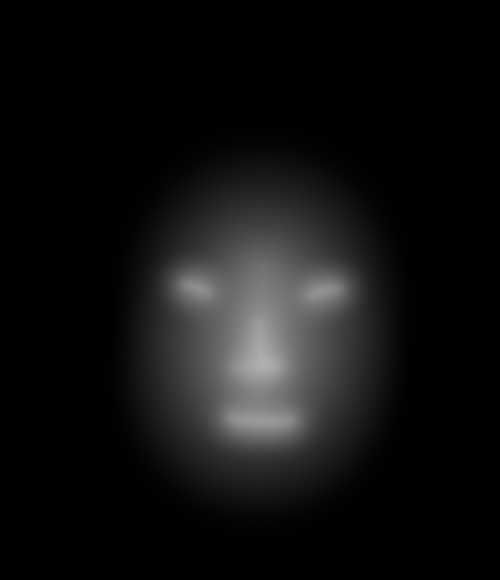}
		\includegraphics[width=0.19\linewidth]{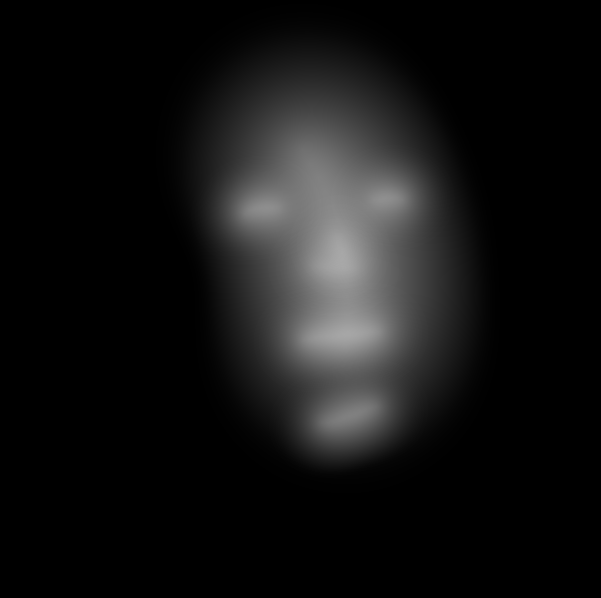}
	\end{subfigure}%
	\vskip\baselineskip
	\begin{subfigure}{\linewidth}
		\label{fig:auto_masks}
		\includegraphics[width=0.19\linewidth]{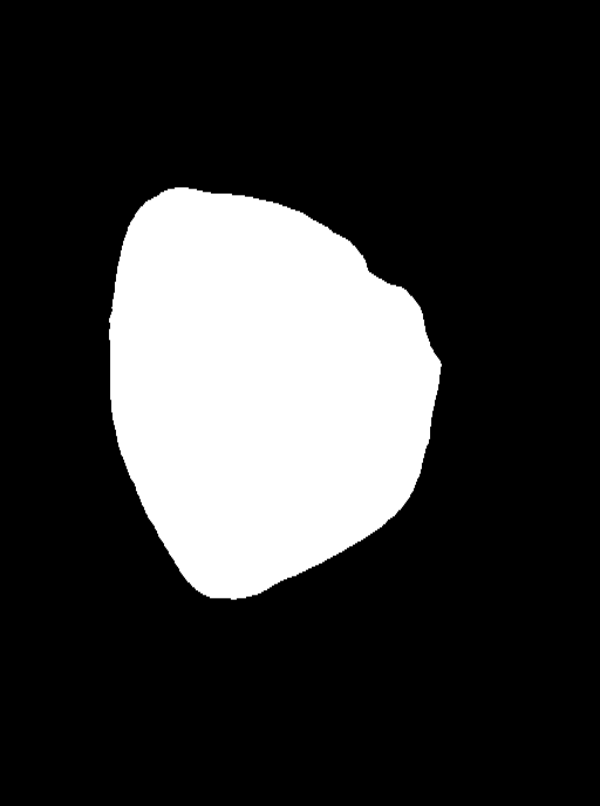}
		\includegraphics[width=0.19\linewidth]{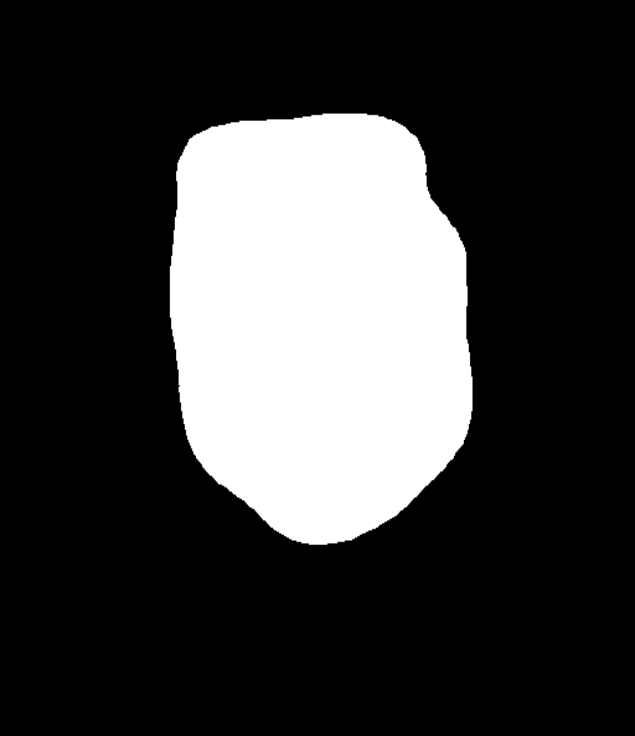}
		\includegraphics[width=0.19\linewidth]{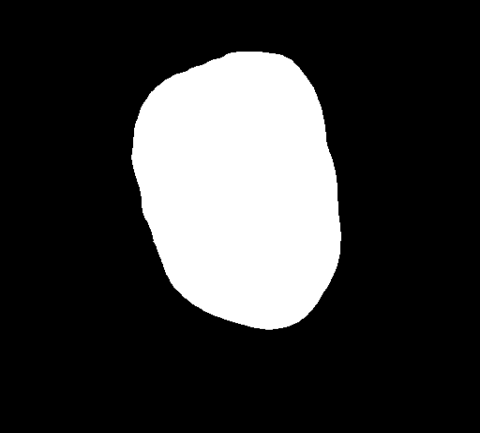}
		\includegraphics[width=0.19\linewidth]{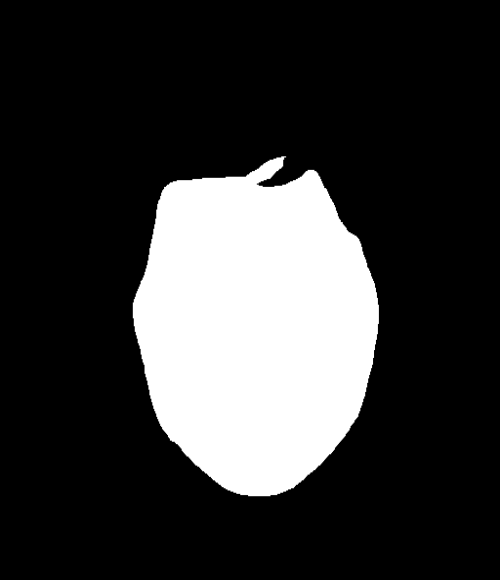}
		\includegraphics[width=0.19\linewidth]{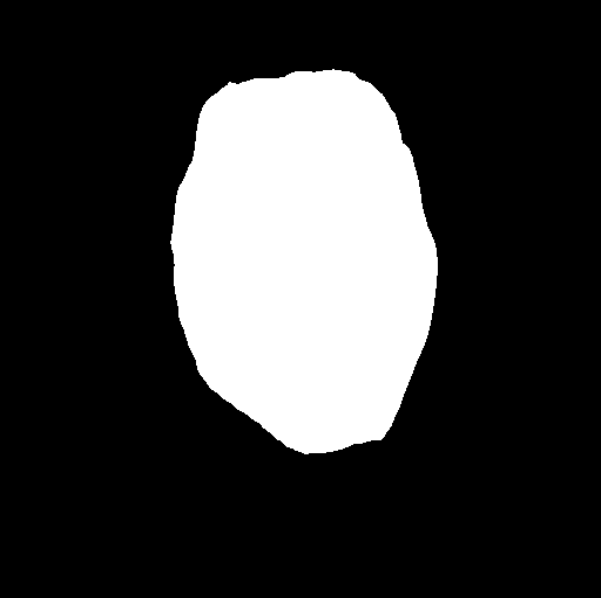}
	\end{subfigure}
	\vspace{1em}
	\caption{Faces and their manual and aitomatic masks.}
	\label{fig:faces_masks}
\end{figure}

Our attack approach is fairly straightforward. Given a source image $I_s$ and target class $C_t$, we paste the face region of $I_t$ into $I_s$ yielding our altered image $I^t_s$. We found this strategy to be very effective in increasing the \conf score, which is to be expected, since the altered image literally comprises the face of another person. Nevertheless, in order to achieve a \stealth of at least $0.5$ and thus a successful attack, the pasting needs to be carefully adjusted. For this we considered the following parameters:

\begin{itemize}
\item \textbf{Position}: The $x$ and $y$ coordinates of the position in the source image, where to place (the center of) the target face.

\item \textbf{Scaling}: The $x$ and $y$ scaling (between $60\%$ and $180\%$) of the pasted target face. 

\item \textbf{Rotation}: The rotation of the pasted target face (between $-40^\circ$ and $40^\circ$, where $0$ represents no rotation).

\item \textbf{Transparency}: In order to mask out non-face regions of the target image, we implemented two approaches, one using manually created masks and one utilizing the face segmentation model BiseNet~\cite{bisnet} to calculate and extracts a mask automatically. 

\begin{itemize}
\item \textbf{Manual masking}: Since there is a limited set of only 10 target face images, we manually constructed continuous valued alpha masks $M_0, ..., M_9$ (see Figure~\ref{fig:faces_masks}) in a simple graphics editor. We highlighted face regions that we believe to be relevant for a face recognition system. Furthermore, for a single attack, a mask $M_i$ is additionally parameterized by a bias $b$ (between $0$ and $1$) and slope $w$ (between $5$ and $40$) resulting in the eventually used mask $M^*_i = \text{sigmoid}((M_i-b)\cdot w)$, where pixel values of $0$ and $1$ represent full transparency and full opaqueness of the respective target image pixel, respectively. This allows gradual tuning of the contour and intensity of relevant regions from the target face that will be inserted into the source image. 
\item \textbf{BiseNet masking}: As an alternative to manual masking, we utilized BiseNet\footnote{\url{https://github.com/zllrunning/face-parsing.PyTorch}} to automatically select the masks. The resulting mask is binary and the transparency is implemented by a gaussian blur of the white pixels with a parameter $\sigma$ (between $0$ and $20$). A $\sigma$ near $0$ is affecting the edge, by increasing the $\sigma$ we increase the effected pixels.
\end{itemize}

\end{itemize}

We use Bayesian Optimization with Gaussian Processes in order to approximate the optimal parameter settings for each one of the 90 attacks. Hereby, the objective to be maximized is:
\begin{equation*}
\text{\conf} + \min(0.5, \text{\stealth})
\end{equation*}
We, therefore, consider \stealth as irrelevant unless it is lower than $0.5$ and thus focus on maximizing \conf while ensuring the attack is successful. 

\begin{figure}
\centering
\begin{subfigure}[b]{0.4\textwidth}
\includegraphics[width=\textwidth,trim={3cm 1cm 3cm 1cm},clip]{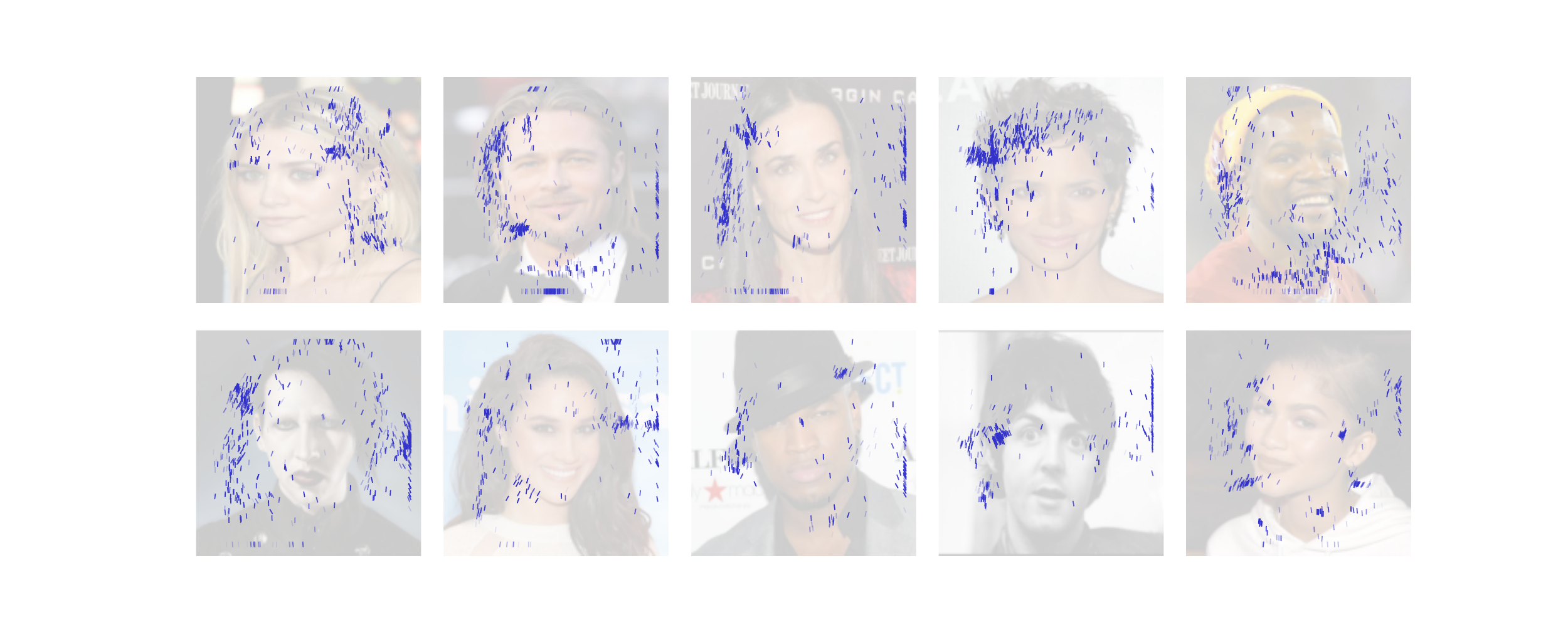}
\end{subfigure}%
\begin{subfigure}[b]{0.06\textwidth}
\includegraphics[width=\textwidth,trim={12.5cm 1cm 0cm 0cm},clip]{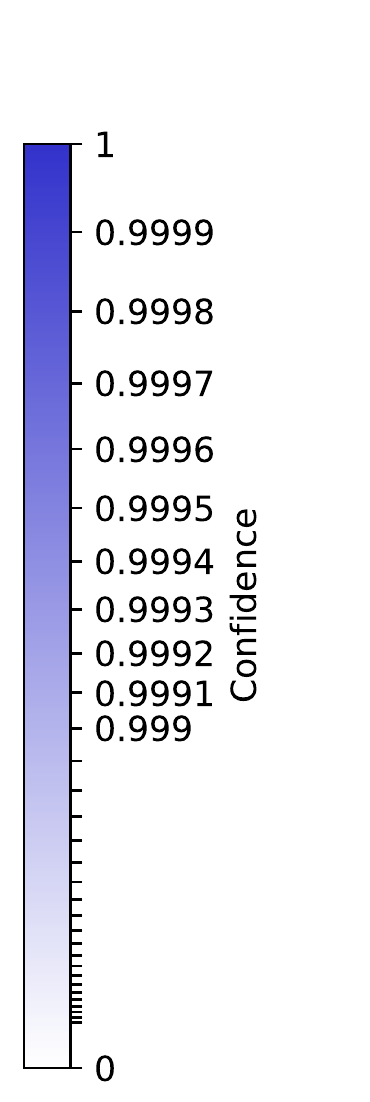}
\end{subfigure}
\caption{Position and rotation (indicated by the line marks, where a vertical line represents no rotation) of various successful query images for the manual masks approach. Their corresponding confidence scores are represented by their opaqueness. Note, that they are power-law normalized, so only very high confidences are visible.}
\label{fig:position_rotation_plot}
\end{figure}

We performed a straightforward ablation study in regards to the parameters and found, that the \textbf{position} is by far the most relevant parameter for both \conf and \stealth. Placing the face over the source face region, effectively erasing most information of the source class, significantly increases the \conf\footnote{Although, we found, that even when placing the target face over the source face, achieving the maximum \conf of $1.0$ is challenging (for example for $s=8$ and $t=2$)}. However, in some cases this also has a negative impact on \stealth, rendering many of attacks as unsuccessful. We found that by favoring positions at the edges of the face or overall source image (allowing about $50\%$ of the target face to be cropped), this problem can easily be overcome, achieving \stealth scores near $1.0$, while still maintaining high \conf. We suspect, that the underlying (SSIM) similarity calculation was improperly calibrated and only considered regions near the source face. 
For the \textbf{scaling} and its considered value range we were not able to find a correlation to the \conf. However a larger scaling lowered the stealthiness, which is to be expected, since the source image is obstructed more. 
In regards to the \textbf{rotation}, we found that a greater angle was slightly disadvantageous for the \conf. Queries of near zero rotation and greater angles near $\pm40^\circ$ achieved on average a confidence of $0.95$ and $0.88$, respectively. 
The effects of position and rotation on \conf is further visualized in Figure~\ref{fig:position_rotation_plot}.

Beside our face pasting attack, we also considered two further attack approaches, namely a face morphing and the more common PGD as a transfer based black box attack (see \hyperref[sec:appendix]{Appendix}).

\section{Evaluation}
\label{sec:eval}
We evaluated our manual and automatic approaches with different parameters for the Bayesian Optimization such as the total number of queries and number of initial queries before utilizing Gaussian Processes. For the other parameters we set the ranges as we considered reasonable (as described in~\ref{sec:attack}).

\begin{figure}
	\centering
	\includegraphics[width=\linewidth]{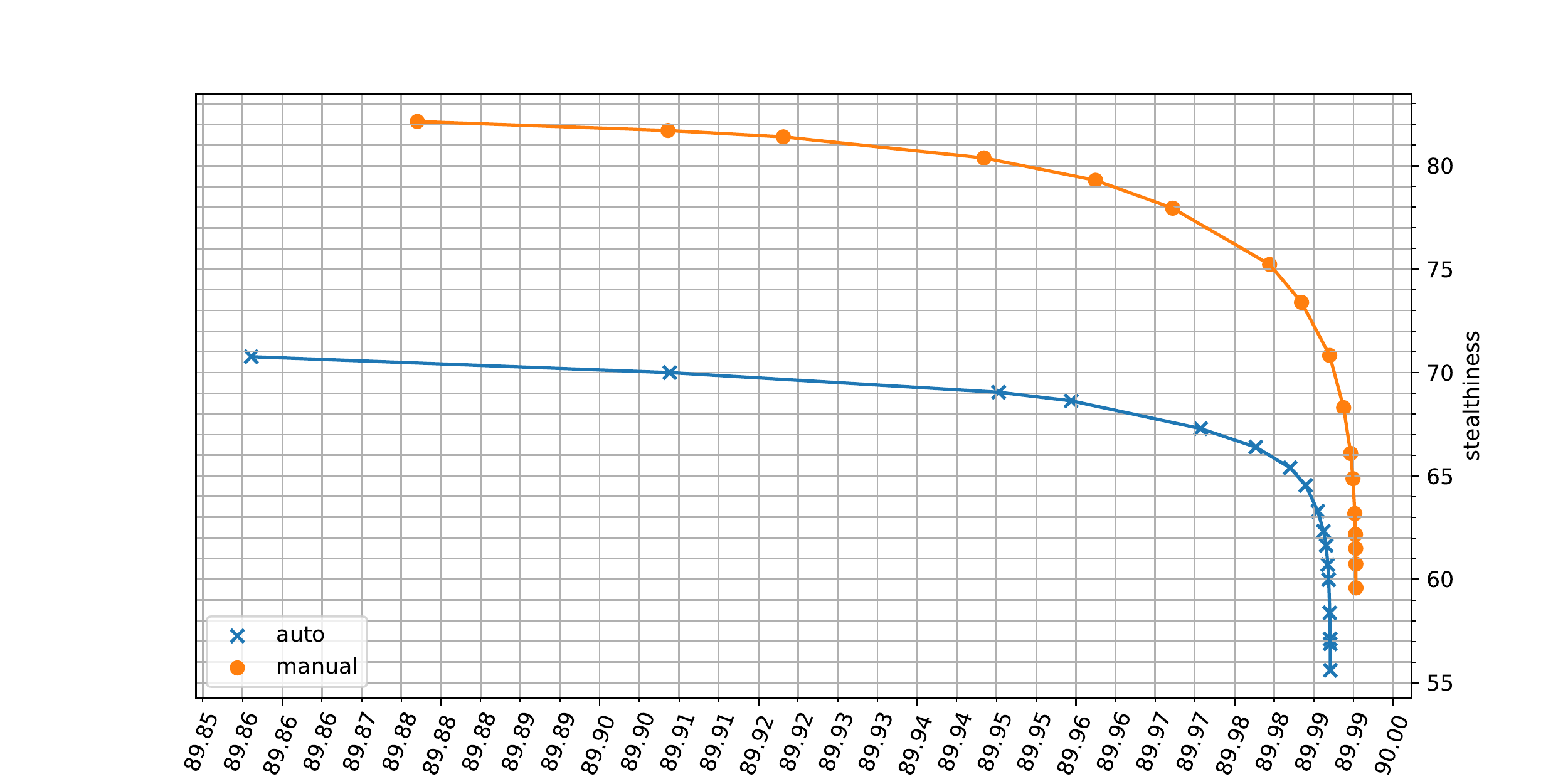}
	\caption{Confidence vs. Stealthiness of manual approach and automatic approach with 200 queries per attack and 50 initial points.}
	\label{fig:conf_stealth}
\end{figure}

We settled with 200 queries in total, comprising 50 initial queries per attack, which results in a \conf of $89.995315$ and \stealth of $58.232452$ for the manual mask approach and a \conf of $89.992083$ and \stealth of $54.825226$ for the automatic BiseNet mask approach. Note, that the overall scores are summed over all attacks. However, we can sacrifice \conf in order to significantly improve \stealth as shown in Figure~\ref{fig:conf_stealth}. In regards to the manual approach, for example, we are able to achieve a \stealth gain of about $20$ while the \conf is only reduced by about $0.03$. In respect to minimum queries for a successful attack, on average $7.689$ and $9.467$ queries are needed for the manual and automatic approach, respectively.
By selecting the best results of both approaches, we arrived at our final submission of  $89.996278$ \conf and $61.19255$ \stealth.

\section{Conclusion \& Future Work}
\label{sec:fw}
The MLSec face recognition competition by Adversa AI and CUJO AI challenged the participants to attack an face recognition black box system in a setting similar to the real world. Our straightforward attack approach pastes target faces into source images and achieves near perfect results in confidence and stealthiness with $200$ queries per attack, which is quite remarkable considering the simplicity. However, the calculation of the stealthiness does not seem to be optimal, since we are able to achieve very high stealthiness for images that are clearly different from the source. This observation shows that every aspect of such systems needs to be carefully designed to ensure an overall robust system.

Future work may comprise more complete evaluation of the parameters for the face pasting and providing insights into the workings of a face recognition model. This competition was designed to reflect a real world scenario, with a  black box system, returning the target confidence score. We should consider other face recognition systems, which return, for example, only the label of the top class.

\section*{Acknowledgement}
This research work has been funded by the German Federal Ministry of Education and Research and the Hessian Ministry of Higher Education, Research, Science and the Arts within their joint support of the National Research Center for Applied Cybersecurity ATHENE.

\bibliographystyle{ACM-Reference-Format}
\bibliography{mybib}
\section*{Appendix}
\label{sec:appendix}

\begin{figure}[b]
	\includegraphics[width=0.2\textwidth]{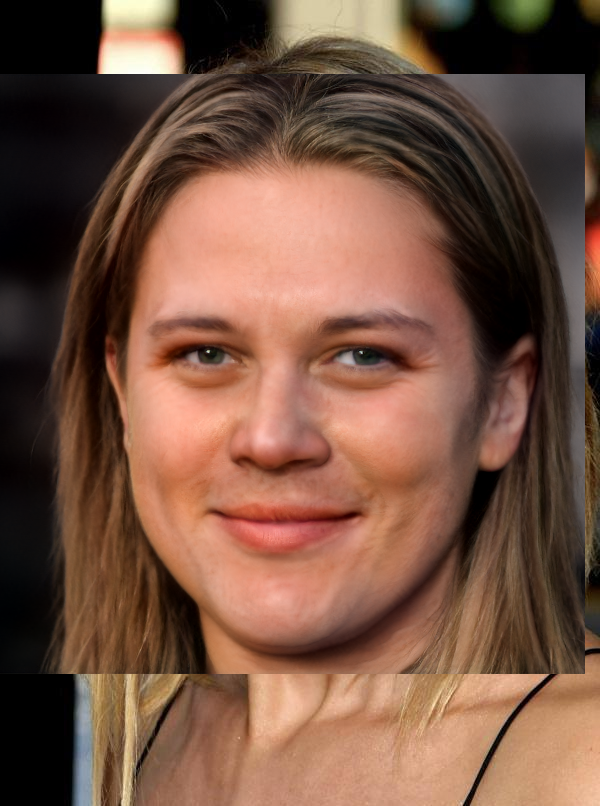}
	\caption{Example of a face morphing attack, for source image $I_0$ and target image $I_1$.}
	\label{fig:stylegan}
\end{figure}

\paragraph{Face Morphing Attack}
We also considered a face morphing attack by interpolating the source and target face in StyleGAN~\cite{stylegan} latent space. However, our initial tests were not promising since the \stealth did not exceed over $0.5$. For example the background, which is left after the face alignment is also interpolated, resulting in many changed pixels and therefore a reduced SSIM and \stealth, as shown in Figure~\ref{fig:stylegan}.

\paragraph{PGD transfer attack}
To evaluate our pasting approach against different attack methods we implemented a PGD transfer attack, where the perturbation is constrained by the SSIM score of the image, such that we reach a \stealth of nearly exact 0.5. We utilized an InceptionResNetV1 pretrained on VGGFace2 as a surrogate\footnote{\url{https://github.com/timesler/facenet-pytorch}}. As the given images are too big for the model and also tend to have a lot of background, we used MTCNN~\cite{Zhang_2016} to detect, crop and scale the face to 160x160 pixels for the model. After the PGD attack we mapped the perturbation back to the original image. But this approach does not lead to a successful attack, with the \conf values reported from the black box system have a mean of $0.0037$ and a maximum of $0.22$.

\end{document}